\documentclass[conference]{IEEEtran}
\IEEEoverridecommandlockouts
\usepackage{cite}
\usepackage{amsmath,amssymb,amsfonts}
\usepackage{algorithmic}
\usepackage{graphicx}
\usepackage{textcomp}
\usepackage{xcolor}

\usepackage{booktabs}
\usepackage{subfigure}
\usepackage{hyperref}

\def\BibTeX{{\rm B\kern-.05em{\sc i\kern-.025em b}\kern-.08em
    T\kern-.1667em\lower.7ex\hbox{E}\kern-.125emX}}
\begin{document}

\title{From 2D Images to 3D Model: Weakly Supervised Multi-View Face Reconstruction with Deep Fusion}
\author{Weiguang Zhao\textsuperscript{1,2$\dagger$}, Chaolong Yang\textsuperscript{1,2$\dagger$}, Jianan Ye\textsuperscript{1,2$\dagger$}, Rui Zhang\textsuperscript{2*}, Yuyao Yan\textsuperscript{2},\\
	 Xi Yang\textsuperscript{2}, Bin Dong\textsuperscript{5},Amir Hussain\textsuperscript{4}, Kaizhu Huang\textsuperscript{3}\thanks{* Corresponding author } \thanks{$\dagger$ Equal contribution}\\
	 \textsuperscript{1}University of Liverpool   $\qquad$  \textsuperscript{2}Xi'an Jiaotong-Liverpool University\\
	\textsuperscript{3}Duke Kunshan University   $\qquad$  \textsuperscript{4}Edinburgh Napier University \\
	\textsuperscript{5}Ricoh Software Research Center (Beijing) Co., Ltd. \\
	{\tt\small \{weiguang.zhao, chaolong.yang, J.Ye13\}@liverpool.ac.uk  \quad \{A.Hussain\}@napier.ac.uk}\\
	{\tt\small \{Bin.Dong\}@srcb.ricoh.com\quad \{kaizhu.huang\}@dukekunshan.edu.cn} \\
	{\tt\small \{Rui.Zhang02, yuyao.yan, xi.yang01\}@xjtlu.edu.cn} 
}

\maketitle

\begin{abstract}
\textbf{While weakly supervised multi-view face reconstruction (MVR) is garnering increased attention, one critical issue still remains open: how to effectively interact and fuse multiple image information to reconstruct high-precision 3D models. In this regard, we propose a novel pipeline called Deep Fusion MVR (DF-MVR) to explore the feature correspondences between multi-view images and reconstruct high-precision 3D faces. Specifically, we present a novel multi-view feature fusion backbone that utilizes face masks to align features from multiple encoders and integrates one multi-layer attention mechanism to enhance feature interaction and fusion, resulting in one unified facial representation. Additionally, we develop one concise face mask mechanism that facilitates multi-view feature fusion and facial reconstruction by identifying common areas and guiding the network's focus on critical facial features (e.g., eyes, brows, nose, and mouth). Experiments on Pixel-Face and Bosphorus datasets indicate the superiority of the proposed method. Without the 3D annotation, DF-MVR  achieves relative  $5.2\%$ and $3.0\%$ RMSE improvement over the existing weakly supervised MVRs, respectively, on Pixel-Face and Bosphorus datasets.  Code has been available publicly at \href{https://github.com/weiguangzhao/DF_MVR}{\tt\small\text{https://github.com/weiguangzhao/DF\_MVR}.}}
\end{abstract}


\section{Introduction}
\label{Intro}

Reconstructing 3D shapes of human faces from 2D images is a challenging yet essential task for numerous applications~\cite{zhang2023domain,yu2023towards,zhao2023fm,liu2023transformer} such as virtual reality, augmented reality, and facial animations. Currently, supervised 3D face reconstruction methods~\cite{fe:1,gu:1} demand a substantial quantity of 3D face meshes or point clouds corresponding to 2D images as ground truth, which are time and/or manpower-consuming.

\begin{figure}[ht]
	\centering
	\includegraphics[width=0.45\textwidth]{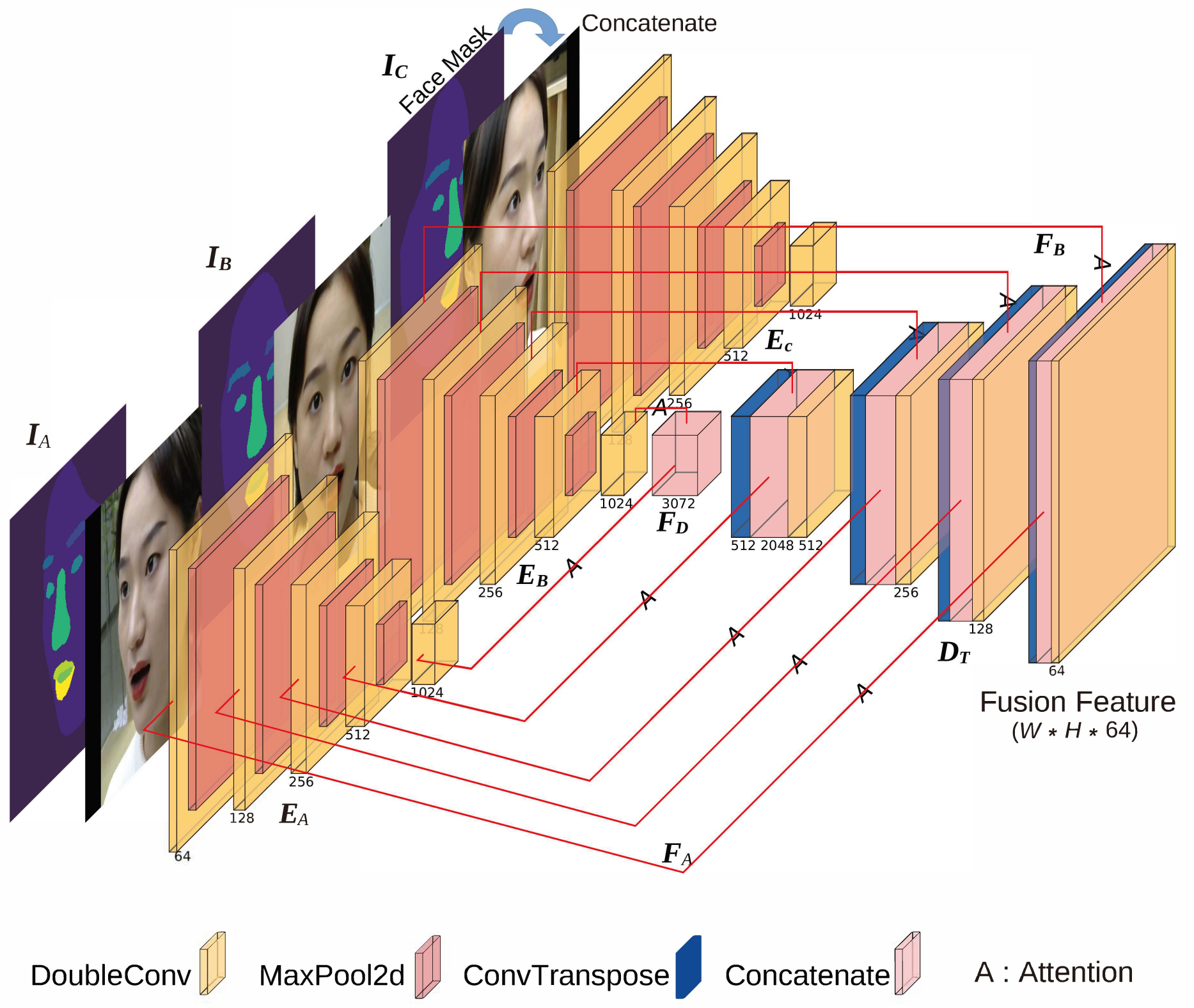} 
	\caption{Facial representation extraction with multi-view face masks and multi-layer attention mechanism. For conciseness, we do not draw the skip connection of the $\mathbf{E}_{C}$, which is similar to the $\mathbf{E}_{A}$.} 
	\label{fig:intro}
\end{figure}

To alleviate the need for 3D face meshes or point cloud data,  recent efforts have shifted to weakly supervised and self-supervised methods~\cite{ba:1,wood20223d,li2023robust}. Most of these methods leverage landmarks and differentiable rendering for training from one single image. Moreover, some weakly supervised multi-view 3D face reconstruction tasks~\cite{de:1,sh:1,xiao2022detailed,kamble20223d,chai2023multi} attempt to reconstruct the more precise face with the geometric constraints contained multiple view images. However, these methods are unfortunately limited because they simply concatenate multi-view image features and overlook the correlations between multi-view features, nor do they pay attention to critical areas (e.g., eye, brow, nose, and mouth) that may impact the reconstruction quality the most.

To cope with these drawbacks, we propose a novel end-to-end weakly supervised multi-view 3D face reconstruction network, namely DF-MVR. Specifically, we utilize the segmentation model~\cite{yu:1} to obtain face masks for the multi-view input images and apply the consistent marking scheme to identify their common regions. This ensures accurate and uniform feature correspondence across different views. As shown in Fig.~\ref{fig:intro},  we design one multi-view feature deep fusion backbone that leverages multi-view face masks to understand the feature correspondences from multi-channel encoders and employs the multi-layer attention mechanism to interact and fuse these features, thereby forming a unified facial information representation. In addition, we propose to develop face masks as the weight map to calculate the pixel-wise photometric loss between rendered images and original images. In this regard, by leveraging the weight map generated from the face mask, we can direct the network to pay greater attention to detailed areas like the eyes, nose, and mouth, enhancing the accuracy of feature extraction.  Based on the proposed multi-view backbone and face mask mechanism, our DF-MVR achieves state-of-the-art performance on both Pixel-Face~\cite{ly:1} and Bosphorus~\cite{SavranADCGSA08} datasets. The contributions of our work are as follows:

\begin{itemize}

        \item  We introduce one novel multi-view feature fusion backbone that utilizes face masks for aligning features from multiple encoders and incorporates a multi-layer attention mechanism to enhance the interaction and fusion of these features, ultimately resulting in a cohesive facial information representation.

        \item  We develop one concise face mask mechanism to facilitate multi-view feature fusion and facial reconstruction. It not only identifies common areas of multi-view images in the fusion progress but also acts as a weight map to encourage the proposed network to pay more attention to critical areas (e.g., eye, brow, nose, and mouth). 
	
	\item On the empirical side, our novel framework achieves $5.2\%$ and $3.0\%$ RMSE improvement over the existing weakly supervised  MVRs, respectively, on Pixel-Face and Bosphorus datasets without 3D annotation.

\end{itemize}

\section{Related Work}

\subsection{Multi-view 3D face reconstruction Methods}
Multi-view methods~\cite{zheng2023multi,peng2024ie,wang2024learning} can provide more geometric constraints and structural information than single-view methods~\cite{zhang2023accurate,zhu2023facescape,chai2023hiface}. There are a few multi-view 3D face reconstruction methods based on weakly supervised machine learning in the literature. Deep3DFace~\cite{de:1} designs two CNN models for predicting 3DMM coefficients and scoring each image. The image with high confidence is used to regress shape coefficients, and the rest images will be used to regress coefficients such as expression and texture. MGCNet~\cite{sh:1} adopts the concept of geometry consistency to design pixel and depth consistency loss. They establish dense pixel correspondences across multi-view input images and introduce the visible maps to account for the self-occlusion. Furthermore, HRN~\cite{lei2023hierarchical}, to be precise, is a semi-supervised approach that utilizes the complete 3D ground truth to generate a deformation and displacement map for training. However, these methods pay less attention to the local features of the face and the feature fusion of multiple images. In contrast, our method not only designs a feature fusion backbone to align and interact features from multiple views but also employs the mask mechanism to promote multi-view feature fusion and focus on the reconstruction of critical areas.

\subsection{Mask in 3D face reconstruction}
Currently, most 3D face reconstruction methods~\cite{de:1,li2023robust} utilize the skin mask~\cite{jones2002statistical} to obtain the pixel positions of the face, thereby eliminating background interference. Deep3DFace~\cite{de:1} proposes a robust, skin-aware photometric loss to reduce the effects of occlusion. FOCUS-MP~\cite{li2023robust} develops a skin mask to attain a highly robust face reconstruction network. While there are some pioneer works probing face masks in face reconstruction, THFNet~\cite{li:2} utilizes face masks to peel off eyes and brows to focus on the texture of the face skin. GFPMG~\cite{zhao2021generative} adopts the face parsing network to remove the occlusion and restore the face image to achieve the full 3D face. Most of these methods utilize the face parsing network for occlusion localization in the task of single-view 3D face reconstruction. Unlike these approaches, our method develops the face mask mechanism, which not only identifies common areas of multi-view images in the fusion progress but also serves as a weight map to constrain the proposed network to focus on critical areas.

\section{Main Methodology}
\label{sec:Method}

\subsection{Overview}
We first provide an overview of our proposed framework, which is shown in Fig.~\ref{fig1}. We decide to exploit three multi-view images of a subject for generating a corresponding 3D face and introduce the \emph{face parse network (a)} to process these three images separately to generate unified standard face masks. An \emph {feature fusion backbone (b)} is designed to fuse the features of multi-view images in deep by sharing a decoder with an attention mechanism to obtain information from the encoder. Moreover, RedNet~\cite{li:1} with involution is adopted as \emph{parametric regression (c)} to regress 3DMM and pose coefficients. The reconstructed 3D face is reoriented utilizing the pose coefficients and then rendered back to 2D. The photo loss between the re-rendered 2D image and the input image at the target view is calculated, while the masks are exploited as the weight map to enhance the backpropagation of the facial features. In this section, we will provide details on each component as below.

\subsection{Face Parse Net}

\begin{figure}[ht]
	\centering
	\includegraphics[width=0.45\textwidth]{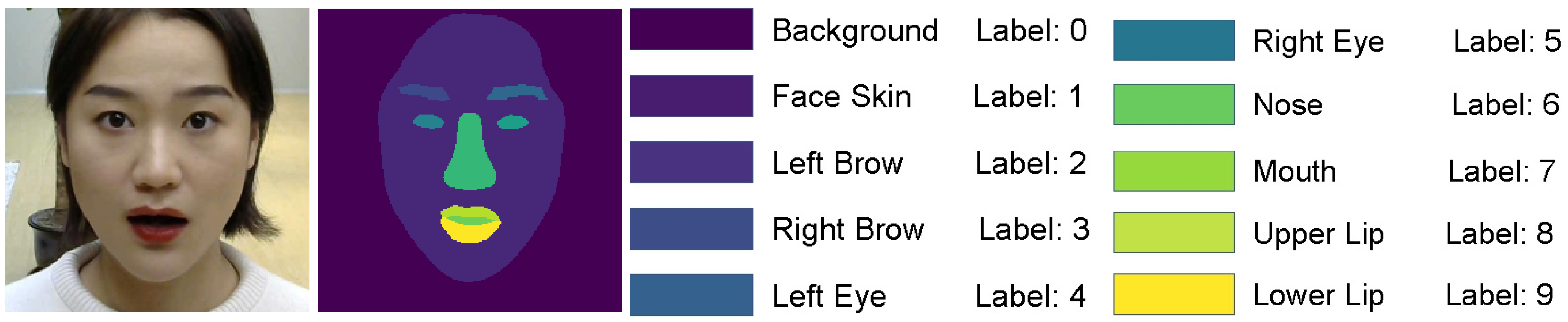} 
	\caption{Face mask annotation.}
	\label{fig2}
\end{figure}

\begin{figure*}[ht]
	\centering
	\includegraphics[width=0.96\textwidth]{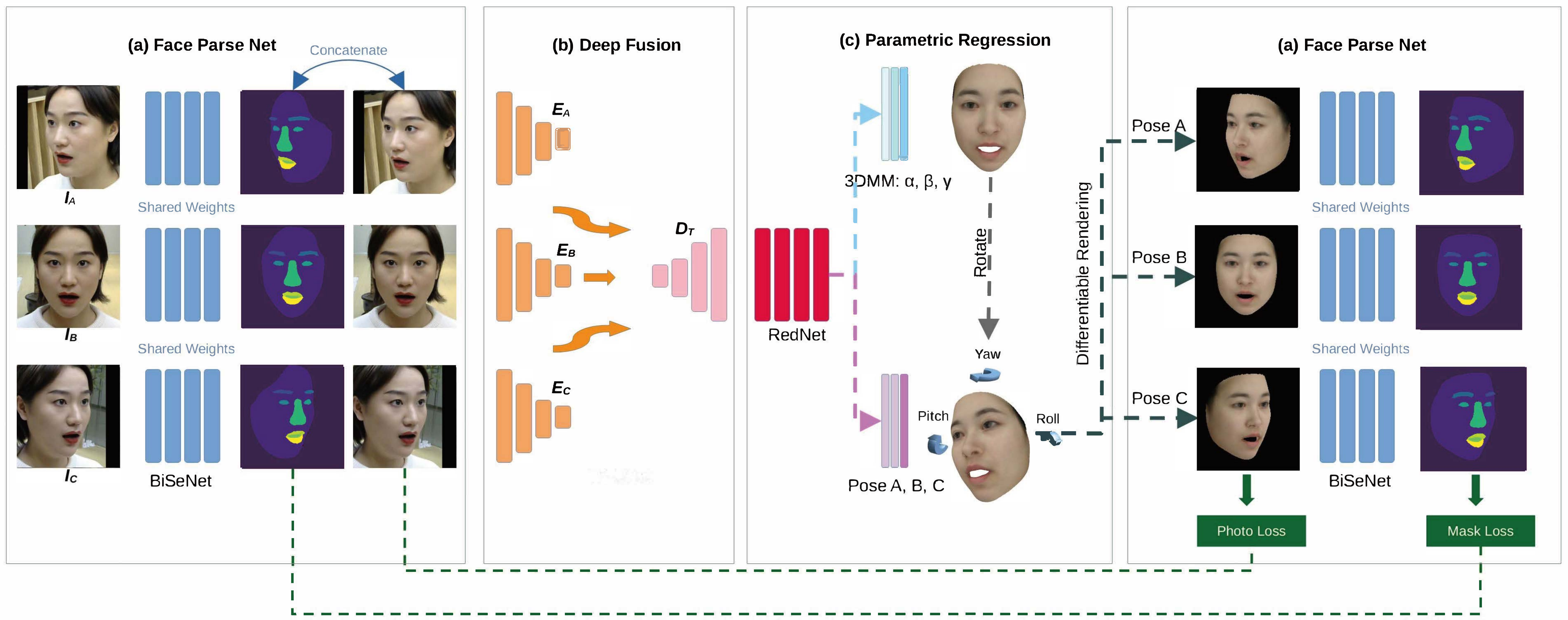} 
	\caption{Overview of DF-MVR}
	\label{fig1}
\end{figure*}

We introduce the face parse network based on BiSeNet~\cite{yu:1} to perform a preliminary analysis of the input image and identify the elements of the image. The generated face mask has only one channel layer.  In order to better highlight the face, excessive elements of face masks, such as hair and neck,  will be removed, and the following parts will be kept: background, face skin, left brow, right brow, left eye, right eye, nose, mouth, upper lip, and lower lip. The reserved parts are marked with different numbers in order to distinguish facial features. As shown in Fig.~\ref{fig2}, one-hot encoding is leveraged to assign the label for these parts. On one hand, the face masks are concatenated with the original images to help the network understand the common area of the multi-view image. On the other hand, the face masks serve as a weight map to calculate the photo loss and mask loss for training.

\subsection{Deep Fusion}
As depicted in Fig.~\ref{fig:intro}, we denote the three-view input images as $\mathbf{I}_{A}$, $\mathbf{I}_{B}$, and $\mathbf{I}_{C}~\in \mathbb{R}^{W \times H \times 3}$, representing the three perspectives of left, front and right. By leveraging pre-trained  BiSeNet, we could attain the corresponding face mask $\mathbf{M}_{A}$,$\mathbf{M}_{B}$, and $\mathbf{M}_{C}~\in \mathbb{R}^{W \times H}$. Corresponding to the input, these three encoders are represented by $\mathbf{E}_{A}$, $\mathbf{E}_{B}$, and $\mathbf{E}_{C}$. The weights of the three encoders are not shared. Encoders are mainly composed of double convolution and maximum pooling. Considering that $\mathbf{I}_{A}$, $\mathbf{I}_{B}$, and $\mathbf{I}_{C}$ actually describe the same object, we only set up one decoder for better-fusing features as well as emphasizing the common features. The decoder $\mathbf{D}_{r}$ is designed to fuse multiple encoder features, mainly composed of ConvTranspose, convolution, concatenate, and skip connection operations.

In this regard, the feature from the $i_{th}$ layer encoder $\mathbf{E}^{i}_{A}$, $\mathbf{E}^{i}_{B}$, and $\mathbf{E}^{i}_{C}$ can be obtained using the following formula:
\begin{small}
\begin{equation}
\begin{aligned}
\mathbf{F}_A^i & =\mathbf{E}_A^i\left(\mathbf{I}_A \oplus \mathbf{M}_A\right) \\
\mathbf{F}_B^i & =\mathbf{E}_B^i\left(\mathbf{I}_B \oplus \mathbf{M}_B\right) \\
\mathbf{F}_C^i & =\mathbf{E}_C^i\left(\mathbf{I}_C \oplus \mathbf{M}_C\right),
\end{aligned}
\end{equation}
\end{small}

where $\oplus$ stands for the concatenate operation. Furthermore, the $i_{th}$ layer decoder feature $\mathbf{D}^{i}_{r}$ could be attained by the following formula:
\begin{small}

\begin{equation}
\begin{aligned}
\mathbf{F}_O^{i-1} & = \mathbf{F}_A^{i-1} \oplus \mathbf{F}_B^{i-1} \oplus \mathbf{F}_C^{i-1} \\
\mathbf{D}_A^i & =\mathcal{A}\left(\mathbf{F}_O^{i-1},\mathbf{F}_A^{i}\right) \\
\mathbf{D}_B^i & =\mathcal{A}\left(\mathbf{F}_O^{i-1},\mathbf{F}_B^{i}\right) \\
\mathbf{D}_C^i & =\mathcal{A}\left(\mathbf{F}_O^{i-1},\mathbf{F}_C^{i}\right) \\
\mathbf{D}_r^i & =\mathbf{D}_A^{i} \oplus \mathbf{D}_B^{i} \oplus \mathbf{D}_C^{i},
\end{aligned}
\end{equation}
\end{small}

where $\mathcal{A}(\cdot)$ donates the attention operation. By following this process, we can obtain the features from each decoder layer sequentially. The feature from the final decoder layer serves as the ultimate facial representation output of the backbone. Finally, the fusion feature size we retain is $224\times224\times64$, in the case where the image size is $224\times224\times3$.

\subsection{Parametric Regression}

We adopt RedNet50 to process the fusion features and regress parameters. RedNet replaces traditional convolution with involution on the ResNet architecture. The inter-channel redundancy within the convolution filter stands out in many deep neural networks,  casting the flexibility of convolution kernels w.r.t. different channels into doubt. Compared with traditional convolution, involution is spatial-specific and able to obtain features on the channel. Therefore, we choose RedNet to perform parameter regression and ablation experiments and verify its effectiveness.
\\
\textbf{3DMM Parameters} regressed in this work include identification, expression, and texture parameters. The 3D face shape $\mathbf{S}$ and the texture $\mathbf{T}$ can be represented as:
\begin{small}
\begin{equation}
	\begin{array}{l}
		\mathbf{S}=\mathbf{S}(\boldsymbol{\alpha}, \boldsymbol{\beta}=\overline{\mathbf{S}}+\mathbf{B}_{i d} \boldsymbol{\alpha}+\mathbf{B}_{exp } \boldsymbol{\beta},\\
		\mathbf{T}=\mathbf{T}(\boldsymbol{\gamma}=\overline{\mathbf{T}}+\mathbf{B}_{t} \boldsymbol{\gamma},
	\end{array}
\end{equation}
\end{small}

where $\overline{\mathbf{S}}$ and $\overline{\mathbf{T}}$ are the average face shape and texture. $\mathbf{B}_{id}$, $\mathbf{B}_{exp}$, $\mathbf{B}_{t}$ are the PCA bases of identity, expression, and texture, respectively. $\boldsymbol{\alpha}$, $\boldsymbol{\beta}$, and $\boldsymbol{\gamma}$ are the parameter vectors that the network needs to regress ($\alpha, \beta \in R^{80}$ and $\gamma \in R^{64}$). By adjusting these three vectors, the shape, expression, and texture of the 3D face can be changed. In order to compare with MGCNet~\cite{sh:1} and  Deep3DFace~\cite{de:1} fairly, we use the same face model. BFM~\cite{pa:1} is adopted for $\overline{\mathbf{S}}, \mathbf{B}_{i d}, \overline{\mathbf{T}}$, and $\mathbf{B}_{t}$.  $\mathbf{B}_{exp}$~\cite{gu:1} is built based on Facewarehouse~\cite{ca:1}.
\\
\textbf{Pose Parameters} are used to adjust the angle and position of the 3D face in the camera coordinate system. We exploit the differentiable rendering~\cite{ra:1} to render the 3D face back to 2D. When the camera coordinates are fixed, we can change the size and angle of the rendered 2D face by adjusting the position of the 3D face in the camera coordinate system. Meanwhile, the position of the 3D face in the camera coordinate system can be determined by predicting the rotation angle and translation in each axis. In order to enhance the geometric constraints of the multi-view reconstruction, we respectively predict the pose of the 3D faces in the multi-view~\cite{dai2023tensorized} instead of only predicting the pose of one view to render 2D images.

\section{Weakly Supervised Training}

\subsection{Photo Loss}
\begin{figure}[ht]
	\centering
	\includegraphics[width=0.40\textwidth]{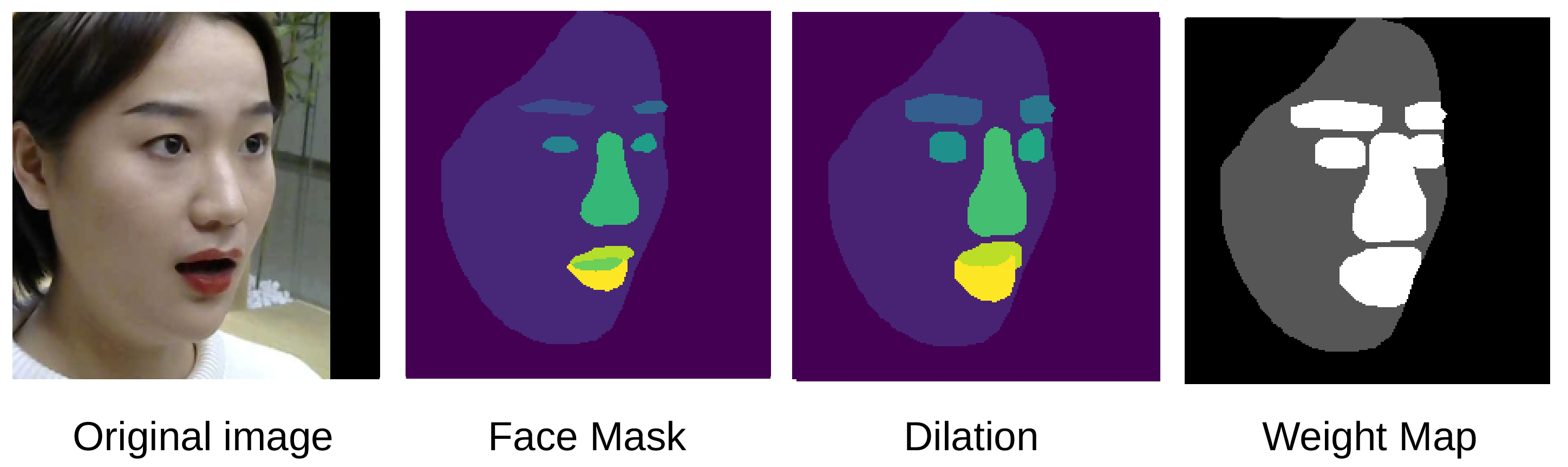} 
	\caption{Weight map}
	\label{fig5}
\end{figure}

Photo loss is often used in weakly supervised face reconstruction tasks~\cite{th:1, te:1}. Distinct from the traditional methods, we impose a weight for each pixel according to the facial features. The weight map is learned by the face mask $\mathcal{M}$ of the original image $I$. In order to enhance the robustness of the weight map, we dilate $\mathcal{M}$ with 20 pixel as $\mathcal{M}_{d}$, as shown in Fig.~\ref{fig5}. The dilated image is divided into three levels to be the weight map. In weight map, facial features are marked as 254, the rest of the facial area is marked as 128, and the background is marked as 32. Our multi-view photo loss can be expressed as:
\begin{small} 
\begin{equation}
	L_{p}=\frac{1}{\mathcal{V}} \sum_{v=1}^{V} \frac{\sum_{i \in \mathcal{P}^{v}} \mathcal{M}_{di}^{v} \cdot\left\|I_{i}^{v}-I_{i}^{v^{\prime}}\right\|_{2}}{\sum_{i \in \mathcal{P}^{v}} \mathcal{M}_{di}^{v}},
\end{equation}
\end{small}
where $\mathcal{V}$ is the number of the reconstructed views. $\mathcal{V}$ is 3 in the proposed model. $\mathcal{P}^{v}$ is the area where the rendered image $I^{v\prime}$ and the original image $I^{v}$ intersect in the current view.  $i$ denotes pixel index, and $\|\cdot\|_{2}$ denotes the L2 norm.

\subsection{Mask Loss}
Photo loss focuses on the pixel difference between two pictures. It is difficult to constrain the size of the facial feature area in the two pictures. For example, the nose color is very similar to that of the cheeks, thereby leading to difficulties for the photo loss to notice the boundary line between them. For this reason, we introduce mask loss to narrow the facial features of the input image and the rendered image. The division and labeling of the facial features are shown in Fig.~\ref{fig2}. According to BiSeNet~\cite{yu:1} we adopt cross entropy to calculate the mask loss $L_{m}$:

\begin{small} 
\begin{equation}
L_{m}=-\frac{1}{N_{p} \mathcal{V}} \sum_{v=1}^{V} \sum_{p=1}^{N_{p}}(y_{p} \log \left(\hat{y_{p}}\right) +\left(1-y_{p}\right) \log \left(1-\hat{y_{p}}\right)),
\end{equation}
\end{small}
where $N_{p}$ is the number of pixels in a face mask. $\hat{y_{p}}$ donates the category prediction of pixels in face mask and $y_{p}$ is the label.

\subsection{Overall Loss}
The overall loss required by our end-to-end net for weakly supervised training can be represented as:
\begin{small} 
\begin{equation}	
L_{all}=\omega_{p} L_{p}+\omega_{m} L_{m} + \omega_{l} L_{l} + \omega_{reg} L_{reg},
\end{equation}
\end{small}
where $\omega_{p}$, $\omega_{m}$, $\omega_{l}$, $\omega_{reg}$ are  the  weights for photo loss$L_{p}$, mask loss$L_{m}$, landmark loss$L_{l}$, and regularization loss$L_{reg}$.  Following Deep3DFace, we set  $\omega_{reg}=3.0\times10^{-4}$. Since $\omega_{2d}$ and $\omega_{3d}$ have been determined, we adjust other parameters as $\omega_{l}=1$,  $\omega_{p}=4$ and $\omega_{m}=3$ respectively. Given that the detailed computation processes of $L_{l}$ and $L_{reg}$ are not central to the main contributions of this paper, we have included them in the supplementary material for completeness.

Note that 3D landmarks are not a must for our method. For general weakly supervised methods, we do not incorporate $L_{l\_3d}$ for comparison~\cite{de:1,sh:1}. Since semi-supervised methods HRN~\cite{lei2023hierarchical} adopt the entire 3D scans to attain training labels, we add $L_{l\_3d}$ for comparative analysis.

\section{Experiment}
\subsection{Experiment settings}
\noindent\textbf{Dataset and Evaluation Metric.} Following BiseNet~\cite{yu:1}, we adopt CelebAMask-HQ~\cite{le:1} to pretrain Face Parse Net. Furthermore, Pixel-Face~\cite{ly:1} and Multi-PIE~\cite{gr:1} are introduced to provide 3D landmarks labels and multi-view faces. Pixel-Face contains $855$ subjects ranging in age from $18$ to $80$ years old. Each subject has $7$ or $23$ samples of different expressions. Pixel-Face has the 3D mesh file of each sample as ground truth but not 3DMM parameters or angle of multi-view images. Additionally, following the previous works, RMSE (mm)~\cite{de:1, sh:1} is used to compute point-to-plane L2 distance between predicted 3D scans and ground-truth 3D scans. 

\subsection{Comparison to SOTAs}

We compare our method with the existing weakly supervised MVRs on both Pixel-Face and Bosphorus datasets. More specifically, only the four methods with codes can be found in the literature related to multi-view weakly supervised 3D face reconstruction. MGCNet~\cite{sh:1} takes multiple images for training and then uses one single image for testing. We select the best results among the three images for display. Deep3DFace~\cite{de:1} does not release the source codes of its scoring network. Therefore, we use their codes to train/test on Pixel-Face to select the best results among the three images.  HRN~\cite{lei2023hierarchical} does not release its training code, but it provides a pre-trained model based on numerous data. To this end, we use its pre-trained model and source code for multi-view face reconstruction. The comparison results have been provided in Tab.~\ref{tab1}. $\dagger$ stands for the necessity to leverage 3D ground-truth (GT) data for semi-supervised  training. HRN$^{\dagger}$ utilizes the entire 3D ground-truth scan to attain both deformation and displacement maps. Ours$^{\dagger}$ only takes a sparse set of 3D landmarks. 

\begin{table}[ht]
 	\caption{Comparison to SOTAs. }
	\centering
    \resizebox{0.40\textwidth}{!}{
	\begin{tabular}{ccc}
	\toprule
		Method                & Dataset     &  RMSE (mm) $\downarrow$              \\ \hline
        Deep3DFace~\cite{de:1}& Pixel-Face  & 1.6641    \\
		MGCNet~\cite{sh:1}    & Pixel-Face  & 1.8877    \\
		Ours                  & Pixel-Face  & \textbf{1.5770}\\   \hline
        Deep3DFace~\cite{de:1}& Bosphorus   & 1.4777    \\
        MGCNet~\cite{sh:1}    & Bosphorus   & 1.4418     \\
		Ours                  & Bosphorus   & \textbf{1.3994} \\  \hline
        HRN$^{\dagger}$~\cite{lei2023hierarchical}   & Pixel-Face  & 1.4061        \\
		Ours$^{\dagger}$      & Pixel-Face  & \textbf{1.4040}         \\ 
        \bottomrule
	\end{tabular}
	}
	\label{tab1}
\end{table}

Given the scarcity of open sources for this task, we put our full codes (training \& test) in supplementary; the pre-trained weight and code will also be provided in our GitHub.

\begin{figure}[ht]
	\centering
	\includegraphics[width=0.40\textwidth]{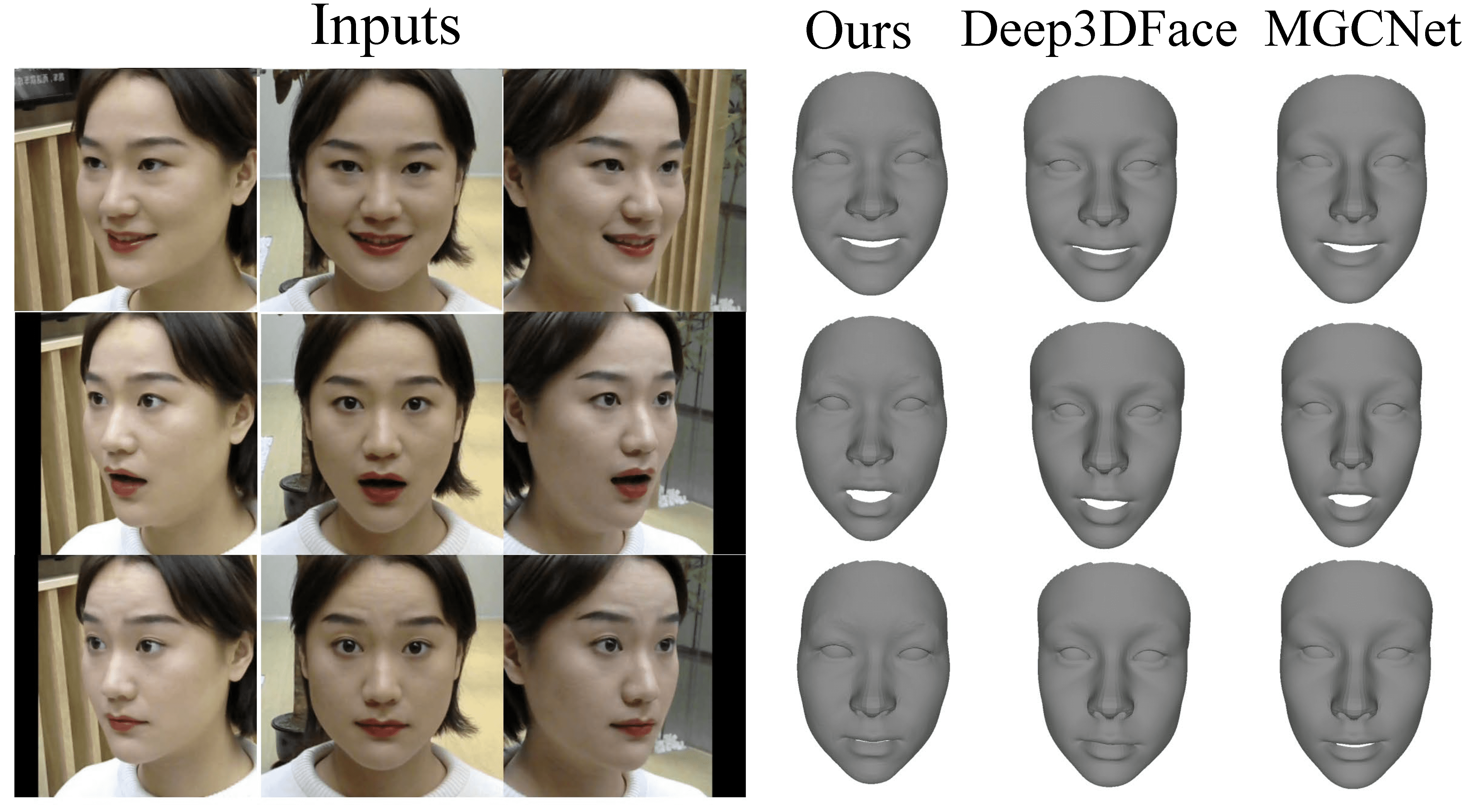} 
	\caption{Qualitative comparison on pixel-face}
	\label{fig6}
\end{figure}

\noindent\textbf{Comparsion on Pixel-Face}. The quantitative results of the comparison are shown in Tab.~\ref{tab1}.  Since Deep3DFace~\cite{de:1} and MGCNet~\cite{sh:1} did not use 3D landmarks, to be fair, we also provide the results of our model without using 3D landmarks for comparison. As observed, our model (without 3D landmarks) shows relative $5.2\%$ improvement compared to the existing methods. While HRN~\cite{lei2023hierarchical} adopts the complete 3D ground-truth scan to attain the deformation map and displacement map, we only take very light annotation, 101 3D landmarks to constraint the face reconstruction. Ours$^{\dagger}$  shows very competitive results with fewer 3D annotation, compared to HRN$^{\dagger}$. The visual comparison is shown in Fig.~\ref{fig6} given 3-view faces. It is evident that our predicted model is more accurate, especially in terms of facial depth estimation in the facial features. 

\begin{figure}[ht]
	\centering
	\includegraphics[width=0.40\textwidth]{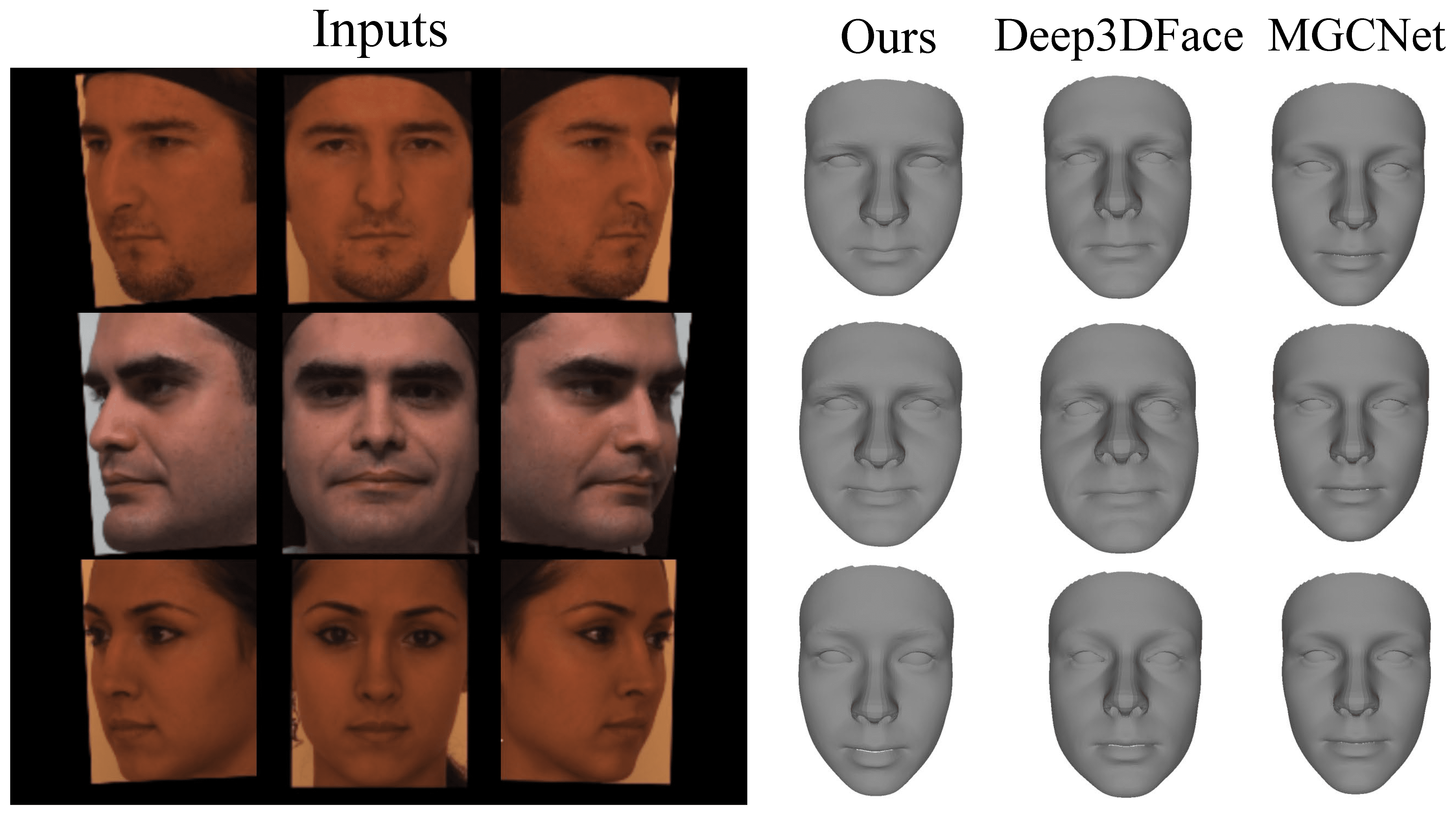} 
	\caption{Qualitative comparison on bosphorus}
	\label{com_bo}
\end{figure}

\noindent\textbf{Comparsion on Bosphorus}. Since the Pixel-Face dataset is collected based on Asian faces, the samples it contains are limited by region. Moreover, the Pixel-Face dataset is only made under one lighting environment and three fixed camera angles. Taking into account the diversity, we adopt the richer Multi-PIE dataset for training and the Bosphorus dataset for testing. Both datasets contain faces across continents and are collected in various lighting environments and shooting angles. Because Multi-PIE does not provide 3D landmarks, we remove $L_{l\_3d}$ to train DF-MVR for this part. The traditional delaunay algorithm is utilized to generate mesh files from point clouds for the Bosphorus dataset. We set the left and right inputs to 45 degrees for comparison. More angle robustness experiments can be later seen in Tab.~\ref{tab4}. The quantitative and qualitative comparisons are shown in Tab.~\ref{tab1} and Fig.~\ref{com_bo}.  Our model (without 3D landmarks) exhibits relative $3.0\%$ improvement compared to the existing methods. 

\subsection{Ablation Study}

In order to verify the effectiveness of the proposed feature fusion backbone and the mask mechanism we designed, we perform more ablation experiments, as shown in Tab.~\ref{tab3}. First, from v1, v2, and v6, it can be found that the multi-view feature fusion network we designed is superior to traditional CNN and Unet in this task. Then, the results of v2 and v6 hint that the multi-layer feature interaction in the feature extraction stage is better than the direct concatenation of features at the end.  Through the RMSE of v3 and v6, it is clear that RedNet performs better than ResNet in this task. For v4, we not only remove the mask loss but also the face mask $\mathbf{I}_{A}$, $\mathbf{I}_{B}$ and $\mathbf{I}_{C}$, which are concatenated to the original image. By comparing v5 and v6,  we can observe that the face mask mechanism promotes the network to generate a higher-precision model. Moreover, we verify the face mask and mask loss separately. Mask loss has a much smaller effect on the accuracy of the model than face mask. While the qualitative result of mask loss is positive for the adjustment of the model during training, we retain this part for reference. Finally, we remove $L_{lan\_3d}$, which means that our model can be trained with only three multi-view images without any 3D label (as denoted as v5). The result also shows that our model is accurate and stable.

\begin{table}[ht]
	\setlength\tabcolsep{3pt} 
	\caption{Ablation Study}
	\centering
        \resizebox{0.46\textwidth}{!}{
	\begin{tabular}{cccccccc}
		\toprule
		Ours & Backbone & /        & $Face mask$ & $L_{l\_3d}$    &  RMSE (mm) $\downarrow$  \\ \hline
		v1   & \textcolor{blue}{CNN}     & RedNet50 & Y         & Y     & 1.5415     \\
		v2   & \textcolor{blue}{Unet}      & RedNet50 & Y         & Y     &  1.5207   \\
		v3   & Ours & \textcolor{blue}{ResNet50}  & Y         & Y      & 1.5206   \\
		v4   & Ours & RedNet50 & \textcolor{blue}{N}         & Y       &  1.4656  \\
		v5   & Ours & RedNet50 & Y         &\textcolor{blue}{N}       & 1.5770    \\
		v6   & Ours & RedNet50 & Y         & Y                 &  \textbf{1.4040}  \\ 
        \bottomrule
	\end{tabular}
         }
	\label{tab3}
\end{table}

\subsection{Robustness Analysis}

\begin{figure}[ht]
	\centering
	\includegraphics[width=0.40\textwidth]{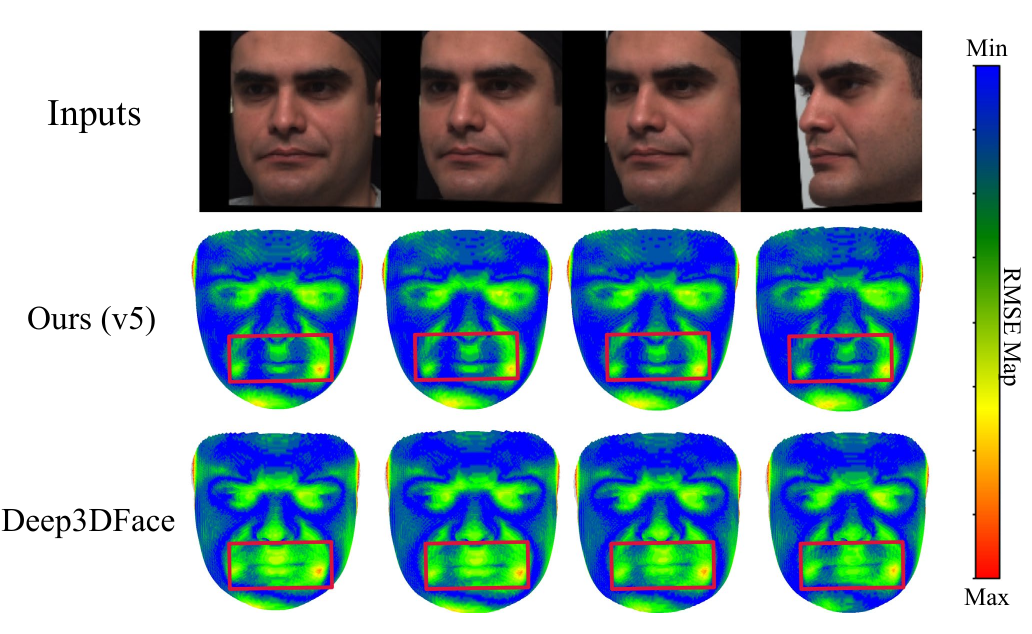} 
	\caption{Robustness analysis.}
	\label{fig_robust}
\end{figure}

Multi-PIE and Bosphorus are collected by different cameras. DF-MVR uses Multi-PIE for training and achieves good test results on Bosphorus, which implies that our model has a certain robustness to camera device changes. Moreover, we further explore the robustness of DF-MVR for shooting angles. We fix the $\mathbf{I}_{B}$ and $\mathbf{I}_{C}$  face angles while adjusting the $\mathbf{I}_{A}$ face angle. Tab.~\ref{tab4} and Fig.~\ref{fig_robust} show that our model consistently achieves the best results in different shooting angles, implying that our model exhibits the best robustness. Random means we choose the degree from $10^{\circ}$, $20^{\circ}$, $30^{\circ}$ and $45^{\circ}$ randomly. The light environment is also random in this experiment.

\begin{table}[ht]
    \setlength\tabcolsep{3pt} 
	\caption{Robustness Analysis on Bosphorus dataset.}
	\centering
	\small
	\begin{tabular}{ccccccllll}
		\toprule
		Method              & $10^{\circ}$    & $20^{\circ}$ & $30^{\circ}$ &$45^{\circ}$ &Random         \\\hline
		Deep3DFace~\cite{de:1}   & 1.5030 & 1.4975    & 1.4924&1.4777 &1.4905\\
        MGCNet~~\cite{sh:1}     &1.4419  &  1.4412  & 1.4413 &1.4418&1.4416 \\
		Ours (v5) &\textbf{1.4251} & \textbf{1.4134}     &\textbf{1.4043} &\textbf{1.3994}  &\textbf{1.4125}\\
        \bottomrule
	\end{tabular}
	\label{tab4}
\end{table}

\section{Conclusion}
In this paper, we design a novel end-to-end weakly supervised multi-view 3D face reconstruction network that exploits multi-view encoding to a single decoding framework with skip connections, able to extract, integrate, and compensate deep features with attention. In addition, we develop a multi-view face parse network to learn, identify, and emphasize the critical common face area. Combining pixel-wise photometric loss, mask loss, and landmark loss, we complete the weakly supervised training. Extensive experiments verify the effectiveness of our model. In future work, we plan to extend our framework by incorporating NeRF and Gaussian Splitting to enhance the representation capabilities of our method.

\bibliographystyle{IEEEbib}
\bibliography{icme2025references}

\end{document}